\documentclass[runningheads]{llncs}
\usepackage[T1]{fontenc}

\usepackage{graphicx}
\usepackage{amsmath}
\usepackage{amssymb}
\usepackage{url}

\begin{document}
\titlerunning{DiffusionXRay}
\title{DiffusionXRay: A Diffusion and GAN-Based Approach for Enhancing Digitally Reconstructed Chest Radiographs}

\author{
  Aryan Goyal\inst{1}\textsuperscript{†} \and
  Ashish Mittal\inst{2}\textsuperscript{†} \and
  Pranav Rao\inst{2} \and
  Manoj Tadepalli\inst{2} \and
  Preetham Putha\inst{2}
}
\authorrunning{A. Goyal, A. Mittal et al.}
\institute{
  Indian Institute of Technology Bombay, India \\
  \email{21d180006@iitb.ac.in}
  \and
  Qure.ai, India \\
  \email{
    ashish.mittal@qure.ai,
    pranav.rao@qure.ai,
    manoj.tadepalli@qure.ai,
    preetham.putha@qure.ai
  }
}

\renewcommand{\thefootnote}{\fnsymbol{footnote}}
\footnotetext[1]{† Equal contribution: Aryan Goyal and Ashish Mittal.}

\maketitle              % typeset the header of the contribution
\begin{abstract}
Deep learning-based automated diagnosis of lung cancer has emerged as a crucial advancement that enables healthcare professionals to detect and initiate treatment earlier. However, these models require extensive training datasets with diverse case-specific properties. High-quality annotated data is particularly challenging to obtain, especially for cases with subtle pulmonary nodules that are difficult to detect even for experienced radiologists. This scarcity of well-labeled datasets can limit model performance and generalization across different patient populations.
Digitally reconstructed radiographs (DRR) using CT-Scan to generate synthetic frontal chest X-rays with artificially inserted lung nodules offers one potential solution. However, this approach suffers from significant image quality degradation, particularly in the form of blurred anatomical features and loss of fine lung field structures.
To overcome this, we introduce DiffusionXRay, a novel image restoration pipeline for Chest X-ray images that synergistically leverages denoising diffusion probabilistic models (DDPMs) and generative adversarial networks (GANs).
% While deep learning models demonstrate impressive capabilities in automated Chest X-ray (CXR) analysis and disease detection, they struggle with low-quality images and require extensive datasets. 
DiffusionXRay incorporates a unique two-stage training process: First, we investigate two independent approaches, DDPM-LQ and GAN-based MUNIT-LQ, to generate low-quality CXRs, addressing the challenge of training data scarcity, posing this as a style transfer problem. Subsequently, we train a DDPM-based model on paired low-quality and high-quality images, enabling it to learn the nuances of X-ray image restoration. Our method demonstrates promising results in enhancing image clarity, contrast, and overall diagnostic value of chest X-rays while preserving subtle yet clinically significant artifacts, validated by both quantitative metrics and expert radiological assessment.

\end{abstract}
\section{Introduction}
%% Para 1 LC detection, modalities, AI and data challenge
Lung cancer, the fifth most common cancer worldwide \cite{sung-2021}, is often diagnosed late due to minimal early symptoms, impacting survival rates \cite{seer}. Although low-dose computed tomography (LDCT) is the preferred screening method for lung cancer \cite{ldct_screening}, chest X-Rays are a widely used alternative due to lower radiation exposure and cost effectiveness. Despite their accessibility, chest X-rays pose challenges in early stage lung cancer detection due to subtle nodules and image variability. Automated nodule detection using deep learning has shown promise in supporting radiologists with early-stage identification in chest X-rays \cite{hendrix-2023}. However, the scarcity of high-quality annotated chest X-ray data remains a major bottleneck in the development of reliable deep learning algorithms.

%% Motivate for synthetic data

To mitigate data scarcity, digitally reconstructed radiographs (DRRs) from CT scans provide a scalable solution \cite{moturu-2018}. However, DRRs suffer from poor contrast, loss of fine structures, and reconstruction artifacts, limiting their diagnostic value. Enhancing DRRs while preserving subtle nodules and fine details is essential for improving lung cancer detection.

% Thus, refining low-quality synthetic X-rays into high-quality images while retaining critical diagnostic details is essential. Such enhancements are vital to ensure that subtle features, indicative of potential nodules or other clinically relevant markers, are preserved, thereby improving the dataset's utility.

% \begin{figure}
% \begin{center}
% \small
% \setlength{\tabcolsep}{2pt}
% \begin{tabular}{ccc}
% {\small Input} & {\small Diffusion-Xray output} & {\small Reference} \\
% {\includegraphics[width=0.2\textwidth]{page1/munit_output.png}} &
% {\includegraphics[width=0.2\textwidth]{page1/diffusionxray_output.png}} &
% {\includegraphics[width=0.2\textwidth]{page1/reference.png}} \\

% % {\includegraphics[width=0.15\textwidth]{figures/imagenet_64x_256x/nn_interp/flower3.png}} &
% % {\includegraphics[width=0.15\textwidth]{figures/imagenet_64x_256x/diffusion_output/flower3.png}} &
% % {\includegraphics[width=0.15\textwidth]{figures/imagenet_64x_256x/reference/flower3.png}} \\
% \\

% \end{tabular}
% \vspace*{-0.2cm}
% \caption{\small Representative DiffusionXray output: 512$\times$512 $\!\rightarrow\!$ 1024$\times$1024 pixels. 
%  \label{fig1}
% }
% \end{center}
% \vspace*{-0.5cm}
% \end{figure}

The lack of extensive datasets containing paired low-quality X-rays (DRRs)\footnote[1]{Throughout this work, 'real low-quality X-rays' refers specifically to Digitally Reconstructed Radiographs (DRRs) obtained via CT projections, while 'generated low-quality X-rays' refer to images synthesized via MUNIT-LQ and DDPM-LQ, designed to mimic DRR degradations.} and high-quality X-ray images is a major limitation in training image enhancement models \cite{lugmayr2019unsupervisedlearningrealworldsuperresolution,Ji_2020_CVPR_Workshops}.  A common approach to generating paired datasets is bicubic interpolation \cite{saharia2021imagesuperresolutioniterativerefinement,weber-2023}, but it fails to replicate the complex degradations present in real low-quality X-rays, leading to poor generalization. Our experimental findings confirm this limitation, highlighting the need for more effective methods to create realistic paired data that accurately capture the unique degradations observed in real low-quality DRRs. To address this,  several approaches have attempted to generate degraded images \cite{peng2024realisticdatagenerationrealworld,yang-2023,zhang2021designingpracticaldegradationmodel}, with some treating the problem as an unsupervised domain adaptation task.

Single-image super-resolution (SISR) is a well-studied topic, with deep learning-based methods ranging from convolutional neural networks (CNNs) to Transformers \cite{zhang-2022,chen-2022} and diffusion models \cite{weber-2023,li-2021}. Recent diffusion models have outperformed GANs in image super-resolution, producing high-fidelity images with fine-grained details \cite{moser2024diffusionmodelsimagesuperresolution}. In the medical domain, however, applying these methods to MRI and X-ray data remains limited.

Our experience with existing models such as SwinIR \cite{liang-2021} showed that while they produce sharp high resolution X-Rays, they often fail to preserve clinically significant details, such as subtle nodules or bone intersections. SwinFIR \cite{zhang-2022} improved global structure retention, but subtle nodules remained faded or distorted. Weber et al. \cite{weber-2023} proposed a super-resolution DDPM trained on the MaChex dataset \cite{weber-2023}, but it focuses on super-resolution of low resolution X-Rays generated from another diffusion model which do not exhibit realistic degradation observed in DRRs.

% \begin{figure}[t]
% \setlength{\tabcolsep}{1.25pt}
% \begin{center}
% \begin{tabular}{ccc}

% \includegraphics[width=0.2\textwidth]{page 2/sample1.png} & 
% \includegraphics[width=0.2\textwidth]{page 2/sample2.png} & 
% \includegraphics[width=0.2\textwidth]{page 2/sample3.png} \\

% \end{tabular}
% \end{center}
% \vspace*{-0.6cm}
% \caption{\small Representative CXRs from the LQ-CXR dataset showcasing subtle pulmonary nodules, synthesized through projection of CT volumes.}
% \vspace*{-0.2cm}
% \label{fig:imagenet_256x_montage}
% \end{figure}

 To address these challenges, We take a novel approach by reframing this problem as a style transfer task—preserving the overall structure while introducing  degradations characteristic of low-quality CXRs. Unlike conventional image processing techniques, which struggle to replicate such degradations, our approach transfers them directly from LQ CXRs to HQ CXRs.

We introduce DiffusionXRay, which employs two independent methods to generate realistic paired data:
(i) MUNIT-LQ, an adaptation of the MUNIT \cite{huang2018multimodalunsupervisedimagetoimagetranslation} framework, and
(ii) DDPM-LQ, a diffusion-based model for learning complex degradations.

% We introduce DiffusionXRay, which explores two independent methods: MUNIT-LQ, an adaptation of the MUNIT \cite{huang2018multimodalunsupervisedimagetoimagetranslation} framework, and DDPM-LQ.

Our proposed method introduces the following novelties:
\begin{enumerate}
    \item To address the lack of accurately paired high- and low-quality training data, we introduce a style transfer approach alongside unsupervised domain transfer. We evaluate two independent methods: (i) MUNIT-LQ (Multimodal Unsupervised Image-to-Image Translation) and (ii) DDPM-LQ, using a two-stage training process.

    \item We train a DDPM on the paired dataset generated using MUNIT-LQ and DDPM-LQ, ensuring realistic degradations and subtle nodules. Both quantitative metrics and expert radiologist evaluations confirm that our model effectively preserves fine anatomical structures, including bronchial details and subtle pulmonary nodules.

    \item We release a dataset of 12,580 low-quality chest X-rays generated from CT projections. Additionally, we provide two low-quality versions of the Chest X-ray \cite{wang2017chestxray} test split (25,596 images), generated using our proposed methods.
    
\end{enumerate}

\section{ Methodology}

\subsection{Datasets}
We introduce LQ-CXR12K (Low-Quality Chest X-ray dataset), comprising 12,580 low-quality chest X-rays generated via projection from low-dose CT scans sourced from our institutional dataset. 

% For training the MUNIT-LQ model, we use unpaired datasets:
% \begin{itemize}
%     \item High-quality (HQ) X-ray dataset: 15,000 high-resolution chest X-rays from our institutional dataset, SubtleNodules-CXR, specifically curated to include solitary subtle nodules.
%     \item Low-quality (LQ) X-ray dataset: LQ-CXR serves as the low-quality component.
% \end{itemize}

The MUNIT-LQ training used unpaired data comprising 15,000 high-quality chest X-ray images from our institutional dataset, SubtleNodules-CXR, specifically selected to contain solitary subtle nodules. For the unpaired LQ component, we employ LQ-CXR12K.

The baseline enhancement model was trained on paired datasets, where the low-quality counterpart was generated using bicubic downsampling on Machex Dataset \cite{weber-2023}. While, the final enhancement model (DDPM-HQ) was trained on the in-house dataset of 300,000 images (HQ-CXR300K), where the degraded images were generated using MUNIT-LQ.

For evaluation, we utilize the ChestX-ray8 \cite{wang2017chestxray} test set containing 25,596 high-quality images. We generate low-quality counterparts using three distinct approaches: bicubic interpolation, MUNIT-LQ, and our DDPM-LQ as illustrated in Fig. \ref{fig:64x_512x_faces}. We release these generated low-quality sets along with their high-quality originals to facilitate future research and benchmarking. 

% We introduce LQ-CXR (Low-Quality Chest X-ray dataset), comprising 12,580 low-quality chest X-rays generated via projection from low-dose CT scans sourced from The Cancer Imaging Archive (TCIA), a publicly available repository of computed tomography (CT) data \cite{the-cancer-imaging-archive-tcia-2024}. 

% The MUNIT-LQ training used unpaired data comprising 15,000 high-quality chest X-ray images from our institutional dataset, SubtleNodules-CXR, specifically selected to contain solitary subtle nodules. For the unpaired LQ component, we employ LQ-CXR.

% The baseline model was trained on paired data, where the low-quality component was generated using bicubic interpolation. While the final enhancement models DDPM-HQ model was trained on an in-house 300,000 images dataset - HQ-CXR, where the degraded image counterparts were generated utilizing MUNIT-LQ.

\begin{figure}
    \centering
    \includegraphics[width=1\linewidth]{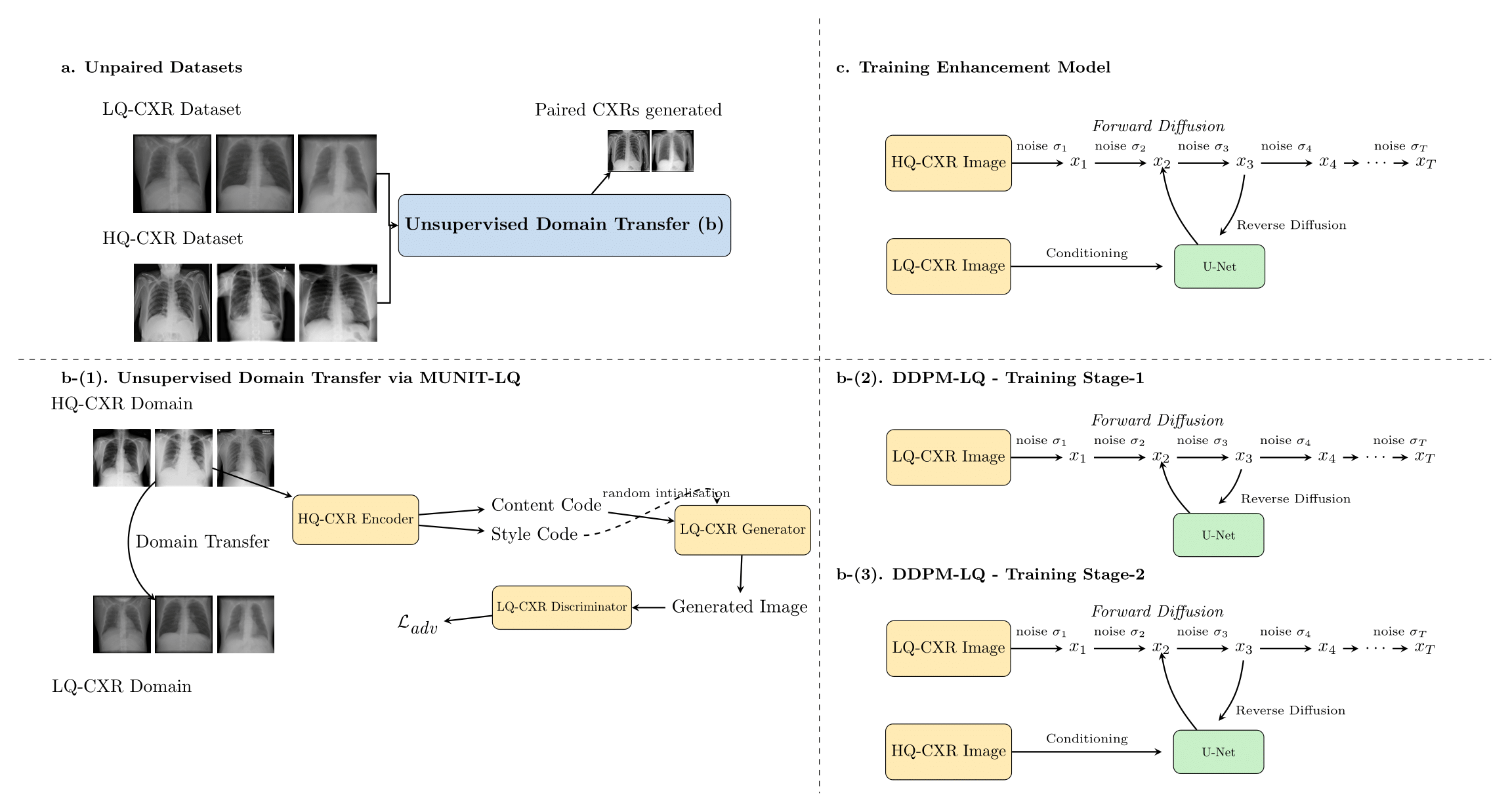}
    \caption{\small Overview of our proposed DiffusionXRay framework. (a) Unpaired high and low-quality (CXRs) serve as training data for our unsupervised domain transfer models. (b) Implementation of domain transfer using: (b-1) MUNIT-LQ for style-guided translation or(b-2) DDPM-LQ for diffusion-based degradation modeling using 2 stage training. (c) The paired data generated from these models is subsequently used to train our final enhancement network which is DDPM-HQ.}
    \label{fig:enter-label}
\end{figure}

\subsection{Synthesizing Chest X-Ray via Projection from CT Scans}
Synthetic frontal chest X-rays can be generated using ray tracing and Beer’s Law applied to CT scans with and without randomly inserted lung nodules, providing a diverse training dataset \cite{Moturu2018CreationOS}.

 However, these digitally reconstructed radiographs (DRRs) suffer from blurred anatomical features and loss of fine lung structures.
 Degradations arise from noise artifacts (e.g., Gaussian noise, beam hardening, streak artifacts, motion distortions) in the CT volume and reconstruction artifacts introduced during the inverse Radon transformation, which converts sinogram data into volumetric images \cite{Benning_Burger_2018}. These distortions hinder the detection of subtle pulmonary abnormalities such as nodules and lesions, as they cannot be accurately simulated using simple degradations like bicubic interpolation or Gaussian filtering.
 
 % The aim of training the final enhancement model is to convert projected X-ray images into refined, high-fidelity data that can serve as reliable training examples
To address this, our pipeline employs MUNIT-LQ and DDPM-LQ to generate realistic low-quality counterparts, enabling the training of an enhancement model that refines DRRs into high-fidelity X-ray images suitable for deep learning applications.

 % This degradation poses a substantial challenge for detecting subtle pulmonary abnormalities such as nodules and lesions, as the loss of detail cannot be simulated via simplistic methods like bicubic interpolation or Gaussian filtering to produce realistic HQ-LQ training pairs. Hence, our pipeline aims to overcome this bottleneck via MUNIT-LR and DDPM-LR for this purpose to generated realistic training data for the main enhancement model.

\subsection{MUNIT-LQ}

The Multimodal Unsupervised Image-to-Image Translation (MUNIT)\cite{huang2018multimodalunsupervisedimagetoimagetranslation} framework implements image translation through disentangled representation via decomposing images into content and style codes. This enables example-guided translation without requiring paired data—making it ideal for generating low-quality chest radiographs (CXRs) from high-quality inputs by leveraging our unpaired HQ-CXR and LQ CXR datasets for training.
The translation process consists of two key steps. First, the content encoder extracts domain-invariant features from the HQ-CXR input images, generating a content code that preserves the essential structural information. Second, this content code is combined with a style code, which can either be randomly sampled from the target domain's style space or derived from a user-provided example image.

% use of 2 gans , loss functions , training and final

 MUNIT-LQ employs two generative adversarial networks: $\text{GAN}_1=\{D_1,G_1\}$ and $\text{GAN}_2=\{D_2,G_2\}$  each trained to reconstruct images within their respective domains i.e high quality CXRs for $\text{GAN}_1$ and low quality CXRs for $\text{GAN}_2$  . $G_1$ is employed to generate low-quality CXRs from the translation stream $\tilde{x}_2^{2\rightarrow 1} = G_1 (z_2 \sim q_2 (z_2|x_2))$. Adversarial training is performed on translated images $\tilde{x}_2^{2\rightarrow 1}$. 

\paragraph{Loss Functions}
The model optimizes a composite loss function comprising adversarial and reconstruction losses. The total loss is defined as:

\begin{equation}
\begin{split}
\mathcal{L} = \mathcal{L}_{x1_{GAN}} + \mathcal{L}_{x2_{GAN}} 
+ \lambda_x(\mathcal{L}_{x1_{recon}} + \mathcal{L}_{x2_{recon}}) \\
\quad + \lambda_c(\mathcal{L}_{c1_{recon}} + \mathcal{L}_{c2_{recon}})
+ \lambda_s(\mathcal{L}_{s1_{recon}} + \mathcal{L}_{s2_{recon}})
\end{split}
\label{eq:total_loss}
\end{equation}

where the image reconstruction loss is computed as:
\begin{equation}
\mathcal{L}_{x1_{recon}} = \mathbb{E}_{x1\sim p(x1)}[\|G_1(E^c_1(x1), E^s_1(x1)) - x1\|_1]
\label{eq:recon_loss}
\end{equation}
The content and style reconstruction losses are given by:
\begin{equation}
\mathcal{L}_{c1_{recon}} = \mathbb{E}_{c1\sim p(c1),s2\sim q(s2)}[\|E^c_2(G_2(c1, s2)) - c1\|_1]
\label{eq:content_loss}
\end{equation}
\begin{equation}
\mathcal{L}_{s2_{recon}} = \mathbb{E}_{c1\sim p(c1),s2\sim q(s2)}[\|E^s_2(G_2(c1, s2)) - s2\|_1]
\label{eq:style_loss}
\end{equation}
The adversarial loss ensures the translated images match the target domain distribution:
\begin{equation}
\mathcal{L}_{x2_{GAN}} = \mathbb{E}_{c1\sim p(c1),s2\sim q(s2)}[\log(1 - D_2(G_2(c1, s2)))] + \mathbb{E}_{x2\sim p(x2)}[\log D_2(x2)]
\label{eq:gan_loss}
\end{equation}
where $\lambda_x$, $\lambda_c$, and $\lambda_s$ are weighting parameters that balance the contribution of each reconstruction term.

In addition to MUNIT’s standard loss functions, we use a domain-invariant perceptual loss via a pre-trained VGG network\cite{johnson2016perceptuallossesrealtimestyle} and a cyclic consistency loss\cite{zhu2020unpairedimagetoimagetranslationusing}.
% While the adversarial loss aids in generating images that match the target distribution, the cyclic consistency loss ensures content preservation across domain translations, stabilizing training.

%WRITE ABOUT LOSS FUNCTIONS
%EXPLAIN GAN VIA EQUATIONS
% ONE DIAGRAM ALSO WITH REPRESENTATION OF OUR DATA

\subsection{DDPM}

% \begin{figure*}[t]
% \vspace*{-.5cm}
% \small
% \begin{minipage}[t]{0.49\textwidth}
% \begin{algorithm}[H]
%   \caption{Training a denoising model $f_\theta$} \label{alg:training}
%   \small
%   \begin{algorithmic}[1]
%     \Repeat
%       \State $(\mathbf{x}, \mathbf{y}_0) \sim p(\mathbf{x}, \mathbf{y})$
%       \State $\gamma \sim p(\gamma)$
%       \State $\bm{\epsilon}\sim\mathcal{N}(\mathbf{0},\mathbf{I})$
%       \State Take a gradient descent step on
%       \Statex $\quad \nabla_\theta \left\lVert f_\theta(\mathbf{x}, \sqrt{\gamma} \mathbf{y}_0 + \sqrt{1-\gamma} \bm{\epsilon}, \gamma) - \bm{\epsilon} \right\rVert_p^p$
%     \Until{converged}
%   \end{algorithmic}
% \end{algorithm}
% \end{minipage}
% \hfill
% \begin{minipage}[t]{0.49\textwidth}
% \begin{algorithm}[H]
%   \caption{Inference in $T$ iterative refinement steps} \label{alg:sampling}
%   \small
%   \begin{algorithmic}[1]
%     \State $\mathbf{y}_T \sim \mathcal{N}(\mathbf{0}, \mathbf{I})$
%     \For{$t = T, \dotsc, 1$}
%       \State $\mathbf{z} \sim \mathcal{N}(\mathbf{0}, \mathbf{I})$ if $t > 1$, else $\mathbf{z} = \mathbf{0}$
%       \State $\mathbf{y}_{t-1} = \frac{1}{\sqrt{\alpha_t}}\left( \mathbf{y}_t - \frac{1-\alpha_t}{\sqrt{1-\gamma_t}} f_\theta(\mathbf{x}, \mathbf{y}_t, \gamma_t) \right) + \sqrt{1 - \alpha_t} \mathbf{z}$
%     \EndFor
%     \State \textbf{return} $\mathbf{y}_0$
%   \end{algorithmic}
% \end{algorithm}
% \end{minipage}
% \vspace*{-0.2cm}
% \caption{Left: Training a denoising model. Right: Inference with $T$ iterative refinement steps.}
% \label{fig:training_inference}
% \end{figure*}

Diffusion Probabilistic Models (DDPMs) outperform GANs in preserving fine anatomical details due to their iterative denoising process and flexible conditioning mechanisms. These properties provide precise control over both the generation process and output characteristics, making DDPMs well-suited for both image enhancement and degradation modelling.
Our image enhancement model and DDPM-LQ are based on the conditional DDPM framework \cite{ho2020denoisingdiffusionprobabilisticmodels}. DDPMs approximate data distributions \( p(x_0) \) through a forward process that progressively adds Gaussian noise according to a variance schedule \( \beta_t \), transforming a data point \( x_0 \) into a sequence of increasingly noisy latents \( x_1, ..., x_T \):
\begin{equation}
q(x_t|x_{t-1}) = \mathcal{N}(x_t; \sqrt{1-\beta_t}x_{t-1}, \beta_t\mathbf{I})
\end{equation}
The reverse process learns to remove noise step-by-step, approximating the true data distribution via a probabilistic model:  
\[
p_\theta(x_{t-1} | x_t) = \mathcal{N}(x_{t-1}; \mu_\theta(x_t, t), \Sigma_\theta(x_t, t))
\]
where the mean \( \mu_\theta \) and variance \( \Sigma_\theta \) are parameterized by a neural network, typically a U-Net, which predicts the noise component \( \epsilon_\theta(x_t, t) \). The reverse process gradually reconstructs the original data by iteratively denoising the latents.

% , and a reverse process modeled using a U-Net architecture \cite{ronneberger-2015}. The forward process gradually transforms a data point $x_0$ into a sequence of increasingly noisy latents $x_1,...,x_T$ through:

For generating low-quality counterparts from high-quality CXRs, DDPM-LQ generates samples from its learned distribution. We initialize our approach using the pre-trained checkpoint from Weber et al. \cite{weber-2023}, which also serves as our baseline. We employ a two-stage training procedure:
\begin{enumerate}
    \item \textbf{Unconditional DDPM Training} – In the first stage, we train DDPM-LQ on 160k real low-quality images -LQ-CXR160K, derived from CT scan projections to learn low-quality characteristics.
    \item \textbf{Conditional Fine-Tuning} – The second stage involves fine-tuning on 300k paired HQ-LQ images, where HQ-CXRs are used as conditioning inputs concatenated with the noise latent during training. The LQ CXR component of the paired dataset was generated using MUNIT-LQ and bicubic downsampling applied on the HQ-CXR300K dataset
\end{enumerate}

This two-stage approach ensures the model learns both general low-quality characteristics and the ability to generate corresponding low-quality images conditioned on high-quality inputs. Finally, our enhancement model is trained on a dataset of 300k paired CXRs, incorporating synthetic low-quality images from DDPM-LQ and MUNIT-LQ.

\section{Experiments}
\subsection{Setup}
 % The primary goal of our experimentation is image enhancement and super-resolution on low-Quality CXRs (DRRs). This is achieved first by generating the realistic low quality counterpart of the given HQ via MUNIT-LQ and DDPM-LQ. 
 The DiffusionXRay training pipeline is essentially a 2-step procedure, the first step being the generation of LQ counterpart given a HQ CXR input, using DDPM-LQ or MUNIT-LQ trained on unpaired datasets: LQ-CXR and HQ-CXR datasets. In the second step we leverage this generated paired dataset to train enhancement models, we employ to train DDPM-based enhancement models. 
 The final enhancement model is then evaluated on the ChestX-ray8\cite{wang2017chestxray} dataset with the low quality X-rays generated using MUNIT-LQ and DDPM-LQ using PNSR and SSIM scores and qualitative evaluation by expert radiologists against the baseline enhancement model.

\subsection{Results}

\begin{figure*}[t]
\vspace*{-0.3cm}
\setlength{\tabcolsep}{2pt}
\begin{center}
\begin{tabular}{cccc}
%{\small NN} &
{\small Input} & {\small Bicubic Baseline} & {\small DiffusionXRay(ours)} & {\small Reference} \\
% {\includegraphics[width=000000.185\textwidth]{figures/faces_64x_512x/nn_interp/84_bb.jpeg}}&
{\includegraphics[width=000000.2\textwidth]{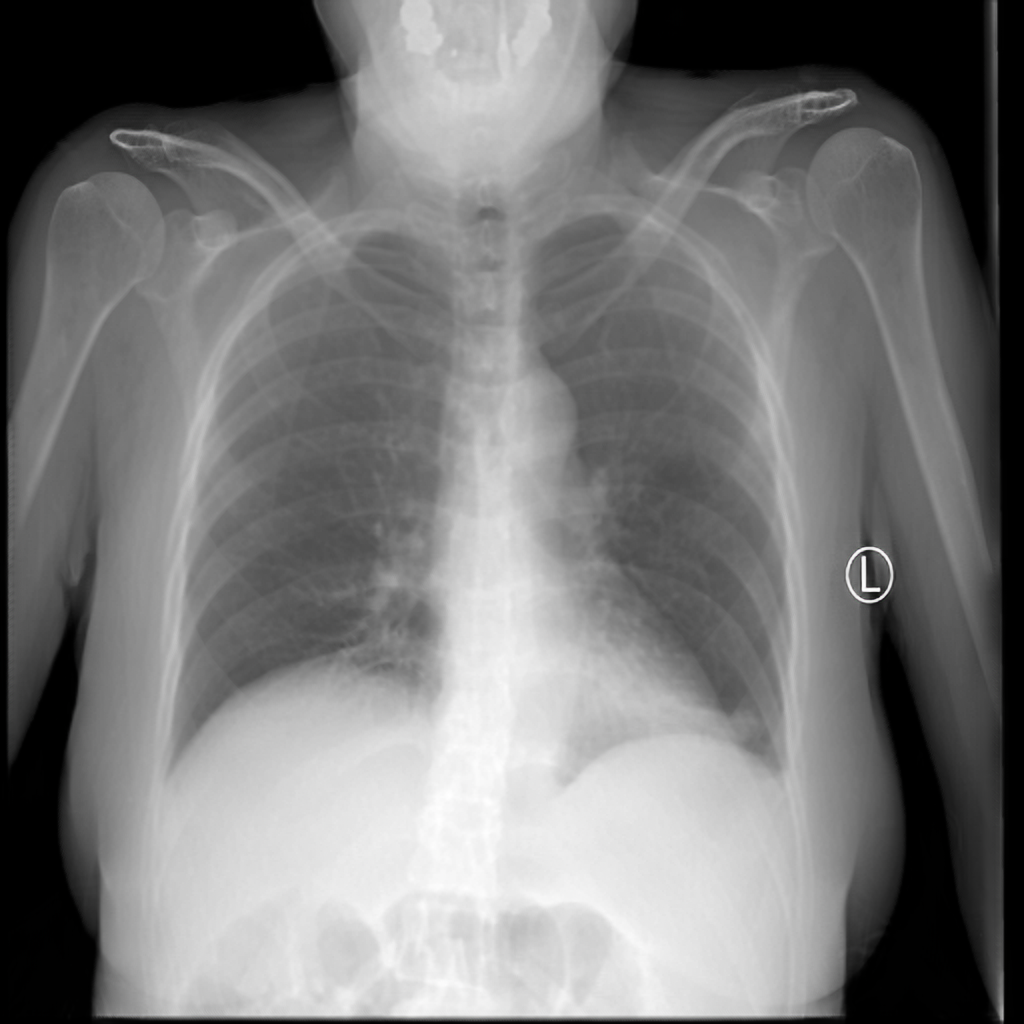}}&
{\includegraphics[width=000000.2\textwidth]{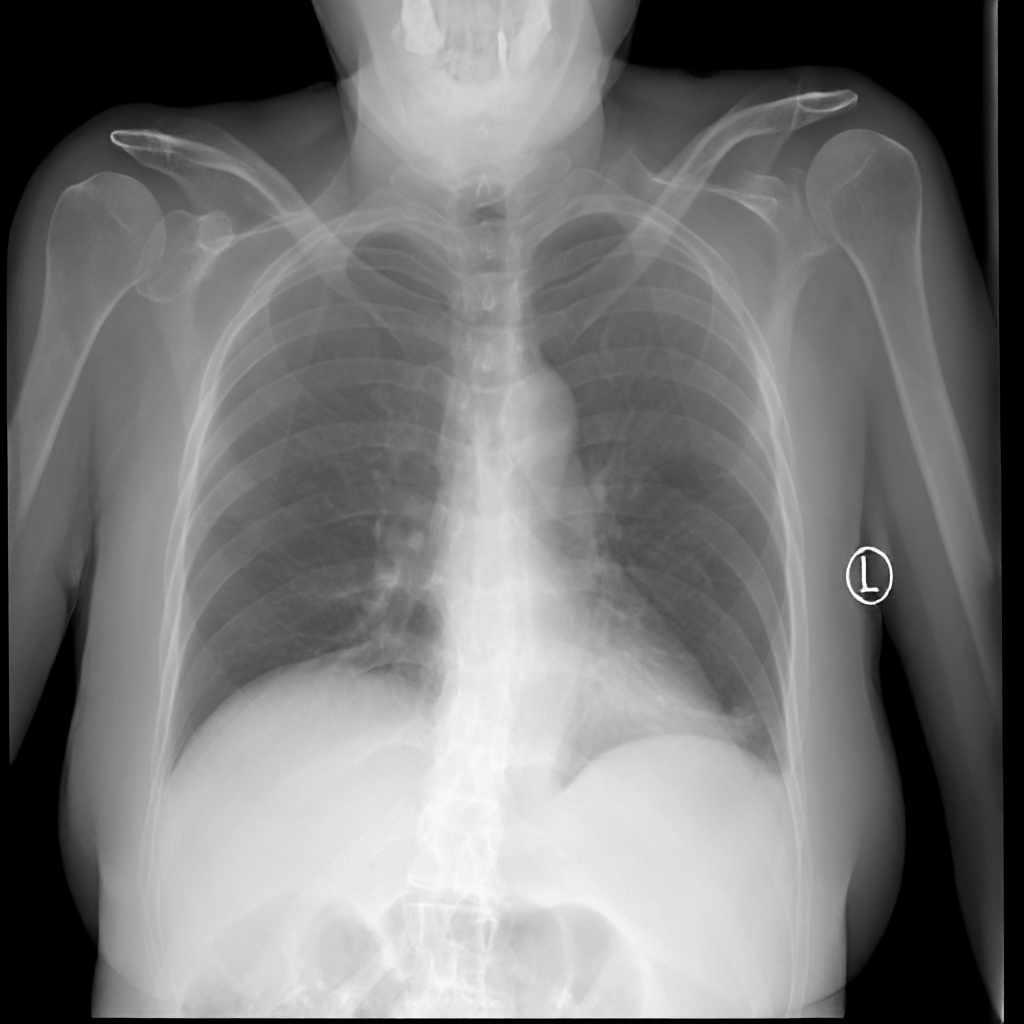}}&
{\includegraphics[width=000000.2\textwidth]{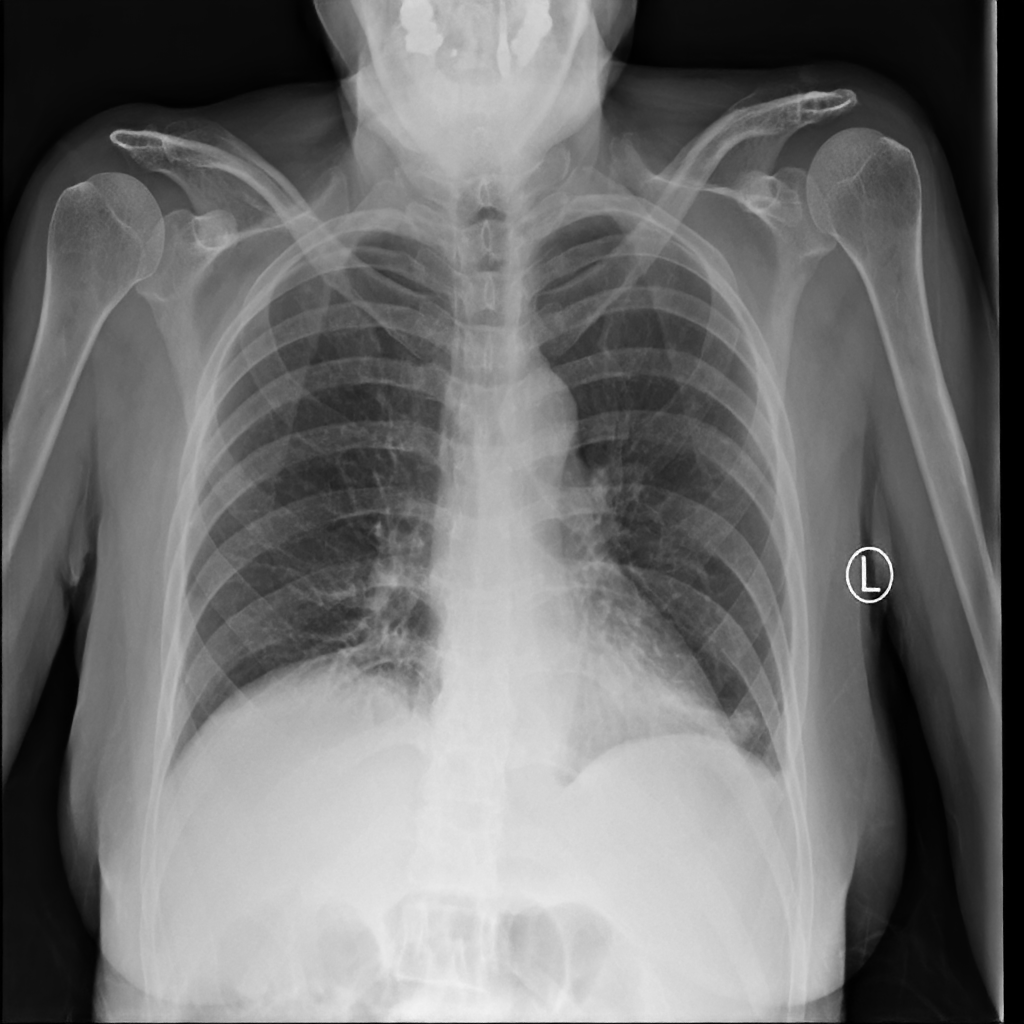}} &
{\includegraphics[width=000000.2\textwidth]{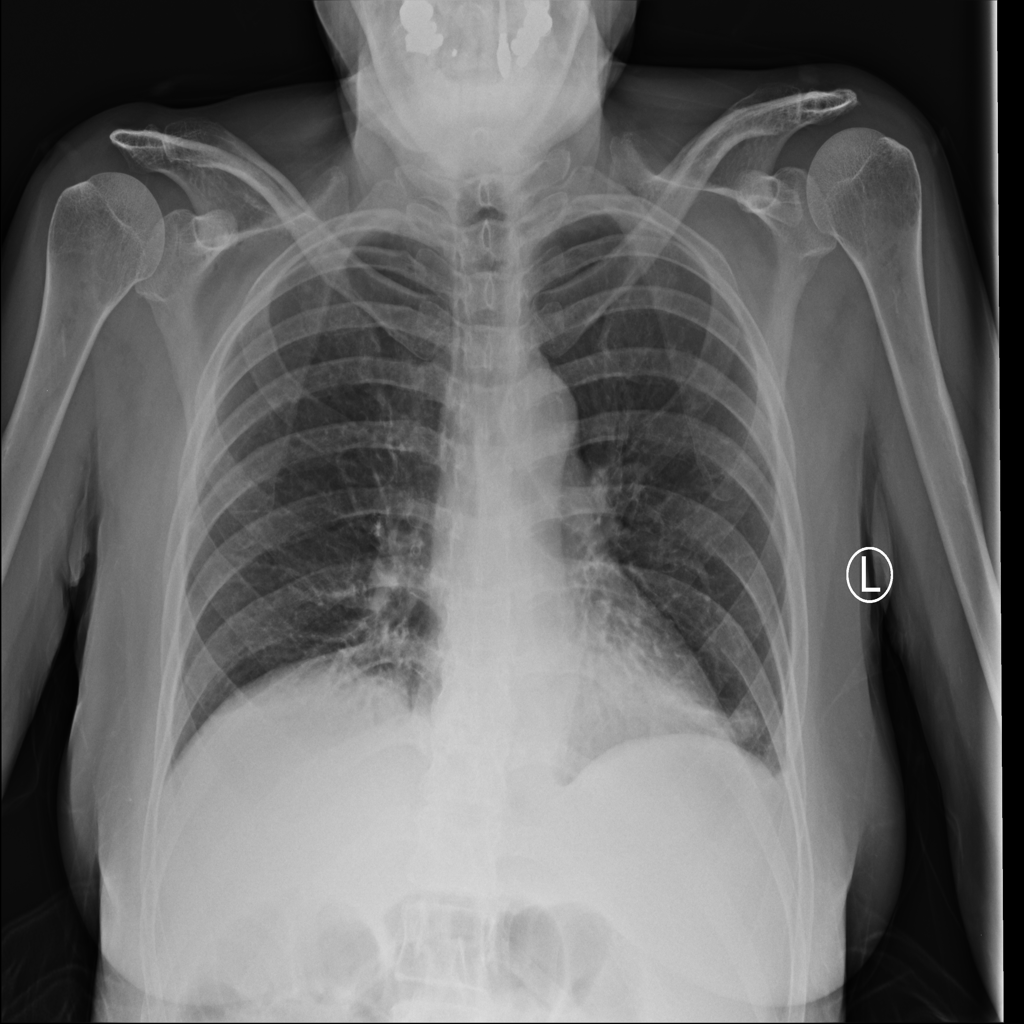}} \\
% {\includegraphics[width=000000.185\textwidth]{figures/faces_64x_512x/nn_interp/84_crop.jpeg}}&
{\includegraphics[width=000000.2\textwidth]{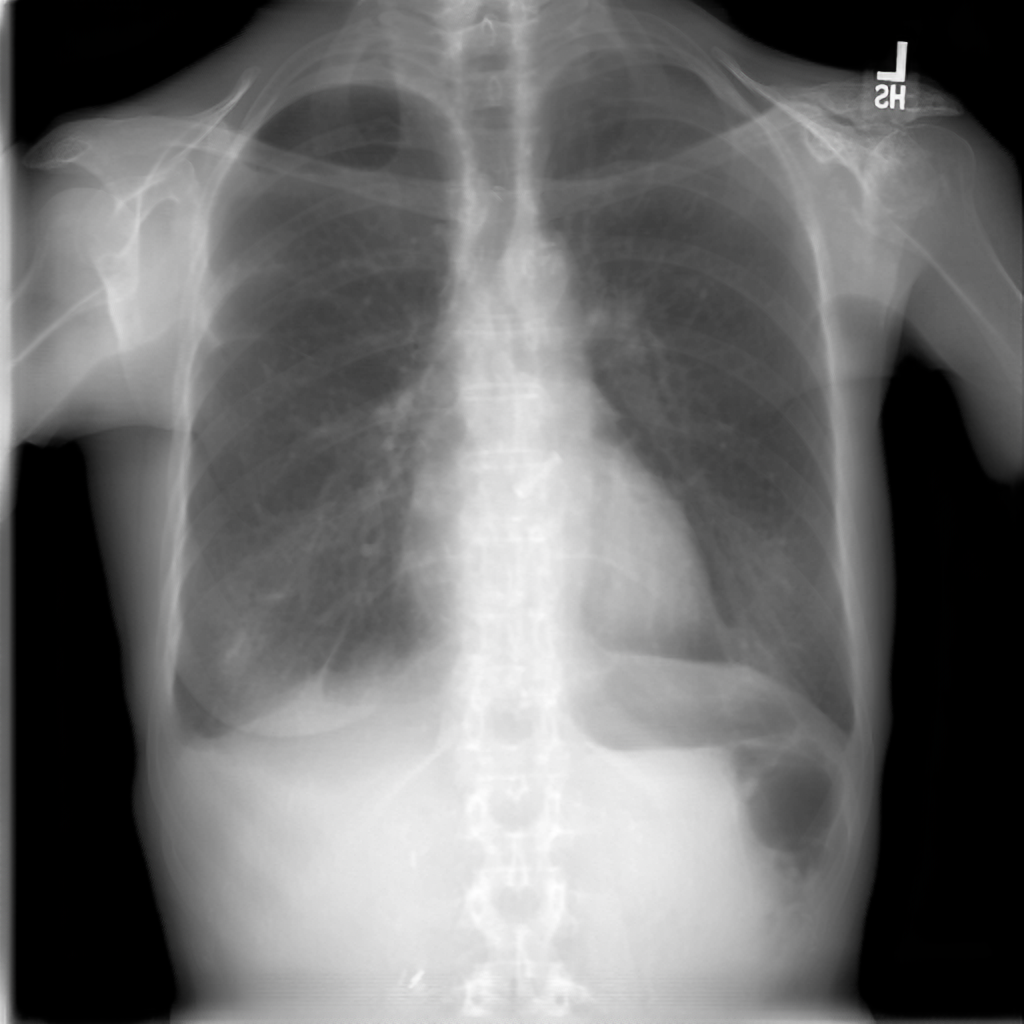}}&
{\includegraphics[width=000000.2\textwidth]{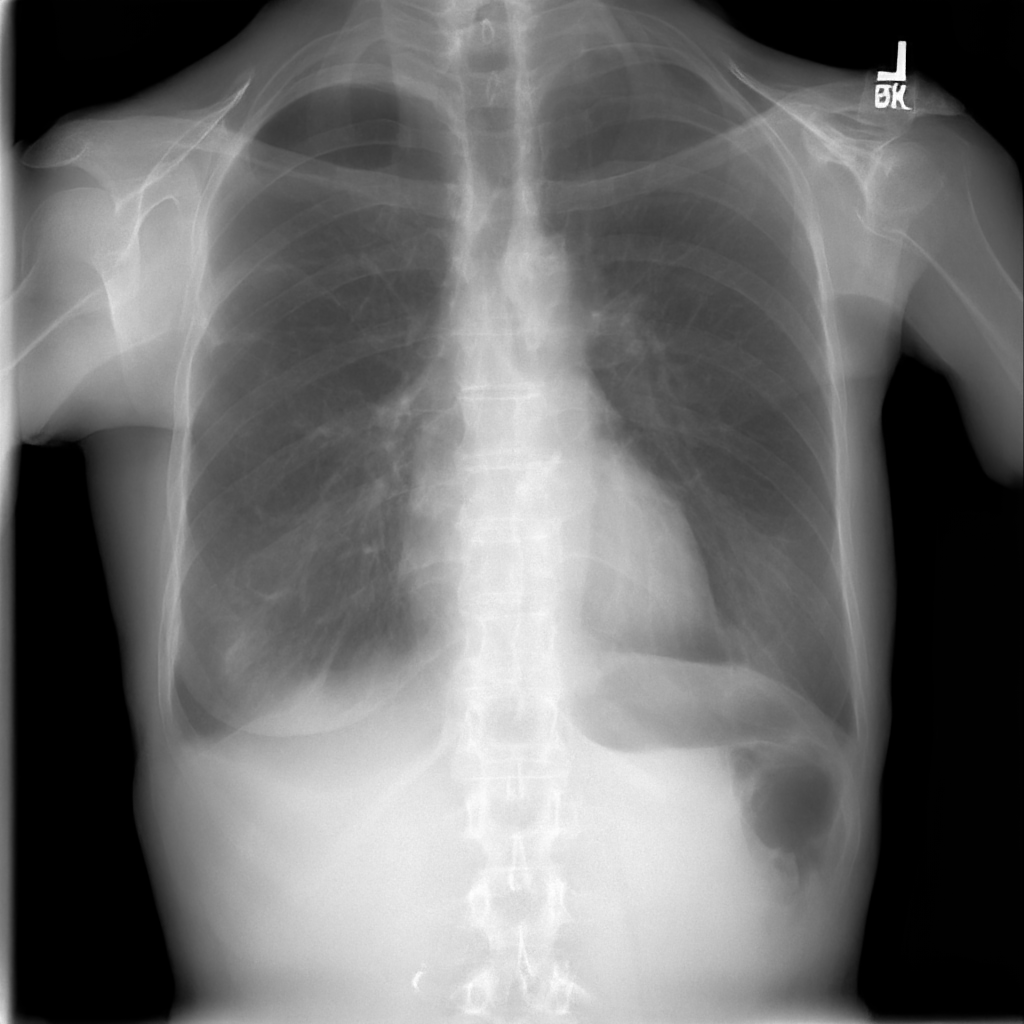}}&
{\includegraphics[width=000000.2\textwidth]{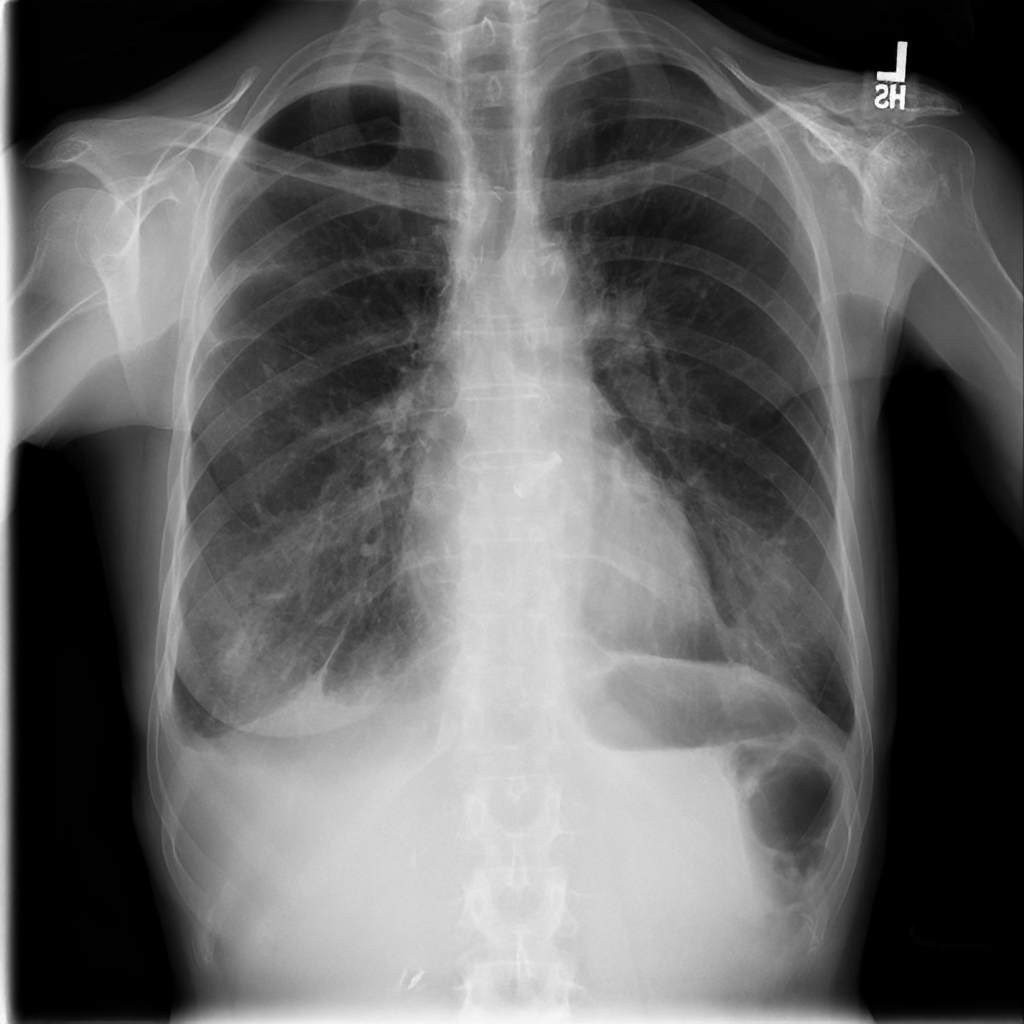}} &
{\includegraphics[width=000000.2\textwidth]{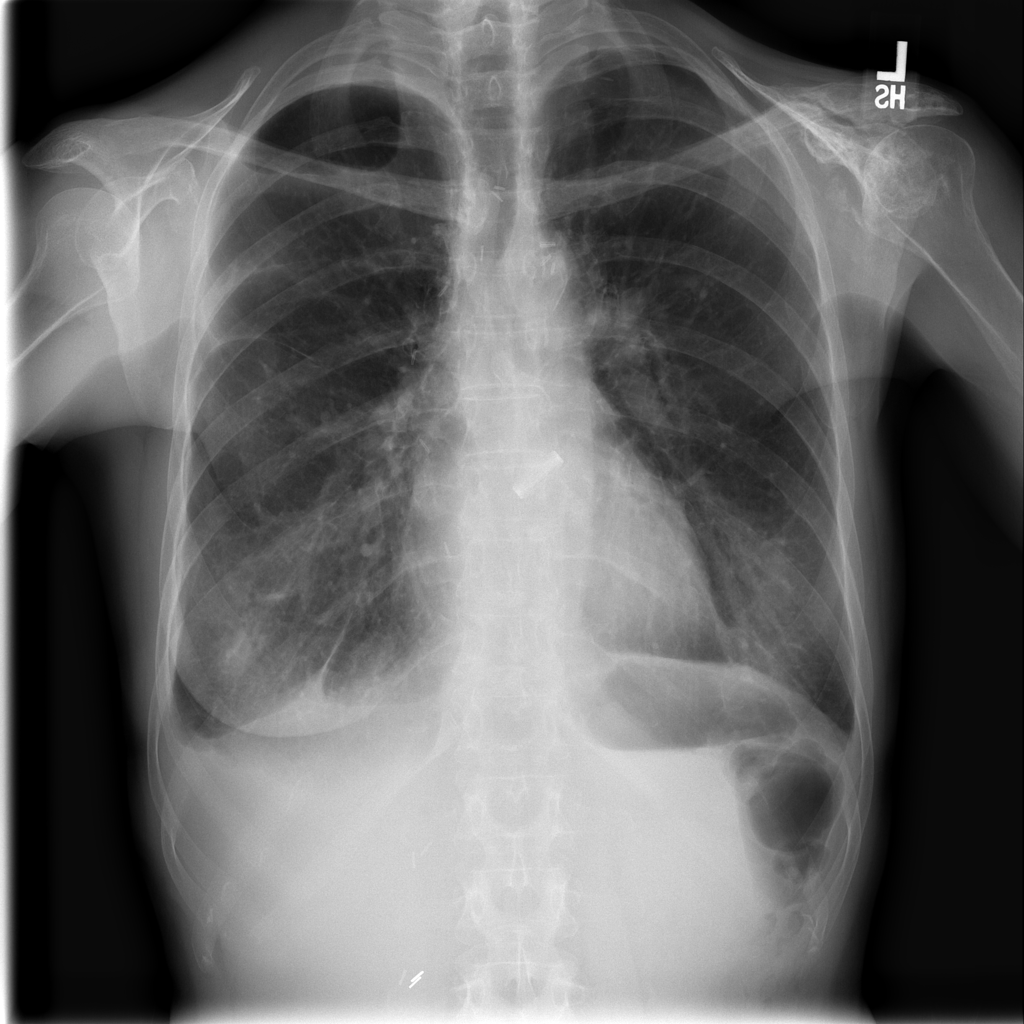}} \\

\end{tabular}
\end{center}
\vspace*{-0.55cm}
\caption{
  \small
  Qualitative comparison of image enhancement and super-resolution results (512$\times$512 $\rightarrow$ 1024$\times$1024) on synthetically degraded chest X-ray (CXR) images. Each input image was a high-quality (HQ) CXR artificially degraded using the MUNIT-LQ framework. The bicubic baseline refers to a DDPM model trained on paired data generated with bicubic downsampling. The reference shown is the original HQ-CXR. \vspace*{-.4cm}
}

\label{fig:64x_512x_faces}
\end{figure*}

\subsubsection{ Automated metrics}

Quantitatively as shown in Table~\ref{tab:psnr_ssim_faces}, DiffusionXRay outperforms the bicubic interpolation low-quality data baseline on PNSR and SSIM. As illustrated in Fig.\ref{fig:64x_512x_faces}, DiffusionXRay produces superior contrast and clarity, whereas the bicubic interpolation baseline DDPM still retains blur and haze artifacts.
\begin{table}[h]
  \caption{\small Quantitative comparison on 512$\times$512 $\rightarrow$ 1024$\times$1024 super-resolution and enhancement with low-quality data generated via MUNIT-LQ and DDPM-LQ on ChestX-ray8\cite{wang2017chestxray} dataset.}
    \centering
    \small
    \scalebox{.95}{    
        \begin{tabular}{|l|c|c|}
            \hline
            \multicolumn{3}{|c|}{\textbf{Results with MUNIT-LQ generated data}} \\
            \hline
            \bfseries Metric & \bfseries Bicubic-DDPM Model & \bfseries Diffusion-Xray (Ours) \\
            \hline
            \textbf{PSNR} $\uparrow$ & 20.08  & \textbf{27.50} \\
            \hline
            \textbf{SSIM} $\uparrow$ & 0.83   & \textbf{0.92} \\
            \hline
            \multicolumn{3}{|c|}{\textbf{Results with DDPM-LQ generated data}} \\
            \hline
            \textbf{PSNR} $\uparrow$ & 19.85  & \textbf{22.21} \\
            \hline
            \textbf{SSIM} $\uparrow$ & 0.78   &  0.78 \\
            \hline
        \end{tabular}
    }
    \vspace*{0.1cm}
  
    \label{tab:psnr_ssim_faces}
\end{table}
\subsubsection{Human Evaluation}

\begin{table}[h!]
\caption{\small Qualitative comparison for nodule reservation and overall quality assessment by expert radiologists for CXR enhancement}
\centering
\begin{tabular}{|l|c|c|}
\hline
      \bfseries Task 1             & \bfseries Q1   $\uparrow$    & \bfseries Q2   $\downarrow$    \\
\hline
DiffusionXRay (ours)              & \textbf{100\%}    & \textbf{0\%}      \\
\hline
Bicubic-DDPM       & 6.6\%    & 30\%     \\
\hline
       \bfseries Task 2           & \bfseries Q1    $\uparrow$   & \bfseries Q2  $\downarrow$     \\
\hline              
DiffusionXRay (ours)             & \textbf{100\%}    & 72.9\%   \\
\hline
Bicubic-DDPM      & 66.7\%   & \textbf{25\%}     \\
\hline
\end{tabular}
\label{tab:nodule-preservation}
\end{table}

Quantitative metrics indicate improvement but fail to adequately capture nodule preservation and clinically significant features. Therefore, we employed radiologist-based mean opinion scores (MOS) to assess nodule visibility and overall image quality, with emphasis on lung field clarity.

In Task 1 (nodule preservation), radiologists reviewed a low-quality input alongside two enhanced outputs—one from bicubic interpolation and one from our model—with nodules highlighted by bounding boxes. They answered:
\begin{enumerate}
    \item Q1 - "Is the nodule easily visible in the post-processed image?"
    \item Q2 - "Could the nodule in the post-processed image be confused with other structures (e.g., bones, artifacts, leads, or buttons)?"
\end{enumerate}
% Q1 - "Is the nodule easily visible in the post-processed image?", Q2 - "Could the nodule in the post-processed image be confused with other structures (e.g., bones, artifacts, leads, or buttons)?"
For Task 2 (quality assessment), radiologists evaluated:
\begin{enumerate}
    \item Q1 - "Has the clarity of the lung fields improved compared to the original projected image?"
    \item Q2 - "Is there a significant increase in noise in the lung fields?"
\end{enumerate}
% For Task 2 (quality assessment), radiologists evaluated: Q1 - "Has the clarity of the lung fields improved compared to the original projected image?", Q2 - "Is there a significant increase in noise in the lung fields?"
Each evaluation used triplets of images, comprising an original CT-projected image and two blinded post-processed versions. As Table \ref{tab:nodule-preservation} demonstrates, our model significantly outperforms the baseline in both tasks, despite introducing some noise patterns, as indicated in Task 2 responses.
\begin{figure*}[t]
\vspace*{-0.3cm}
\setlength{\tabcolsep}{2pt}
\begin{center}
\begin{tabular}{cccc}
%{\small NN} &
{\small Input} & {\small Bicubic} & {\small MUNIT-LQ} & {\small DDPM-LQ (ours)} \\

% Row 2
{\includegraphics[width=0.2\textwidth]{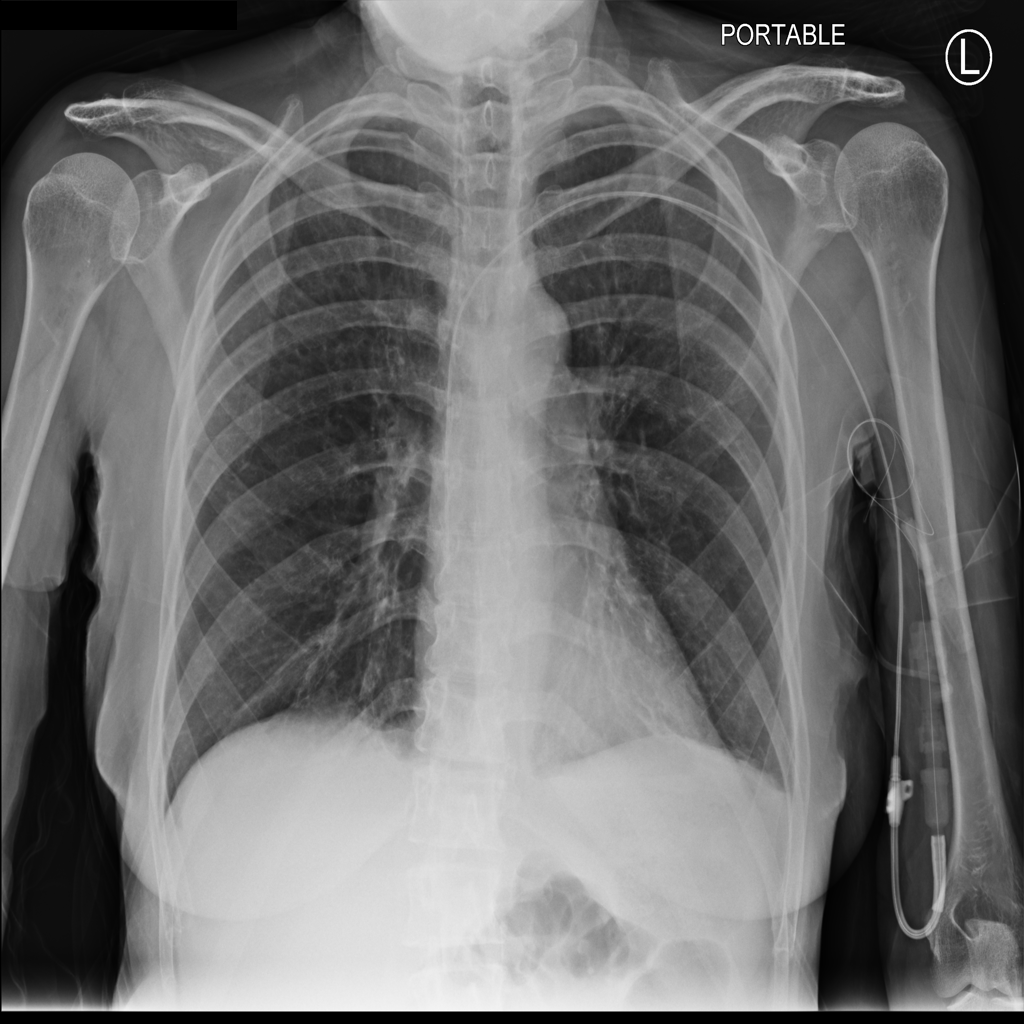}}&
{\includegraphics[width=0.2\textwidth]{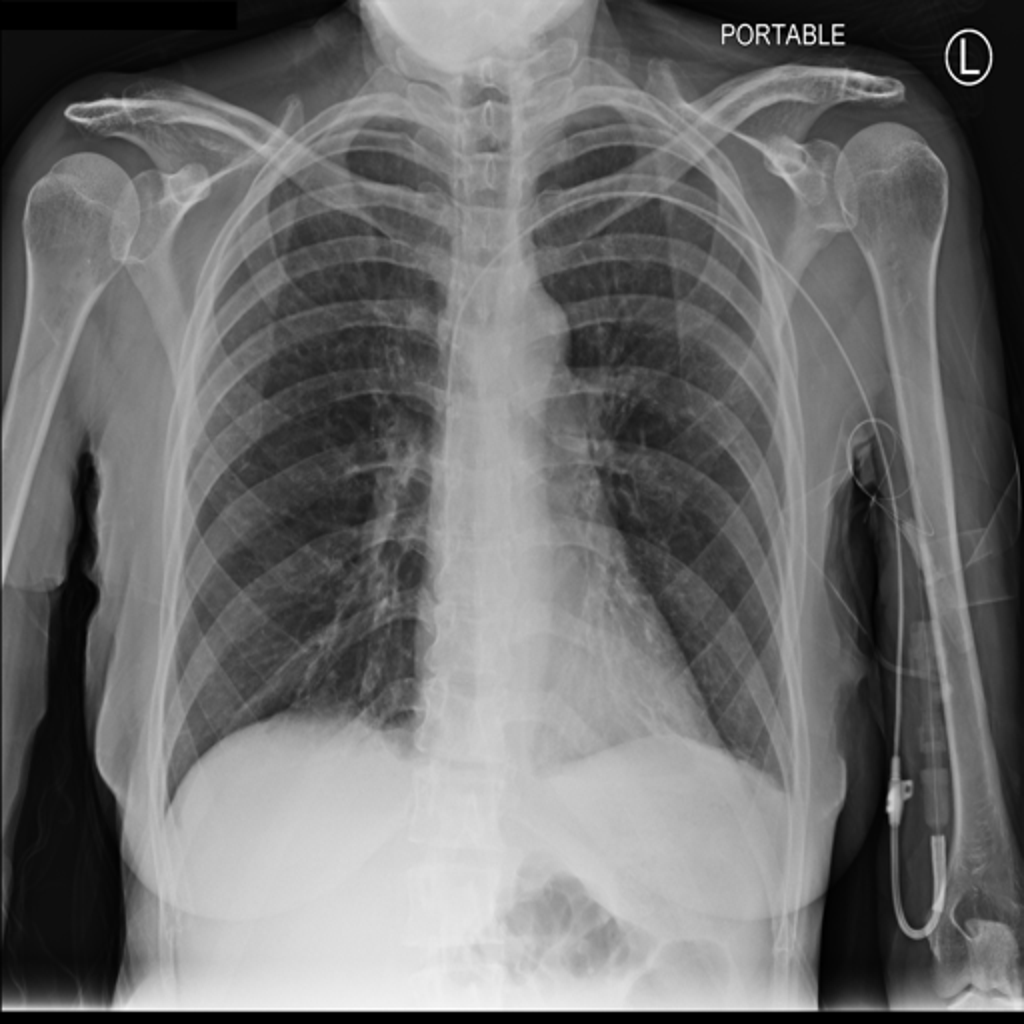}}&
{\includegraphics[width=0.2\textwidth]{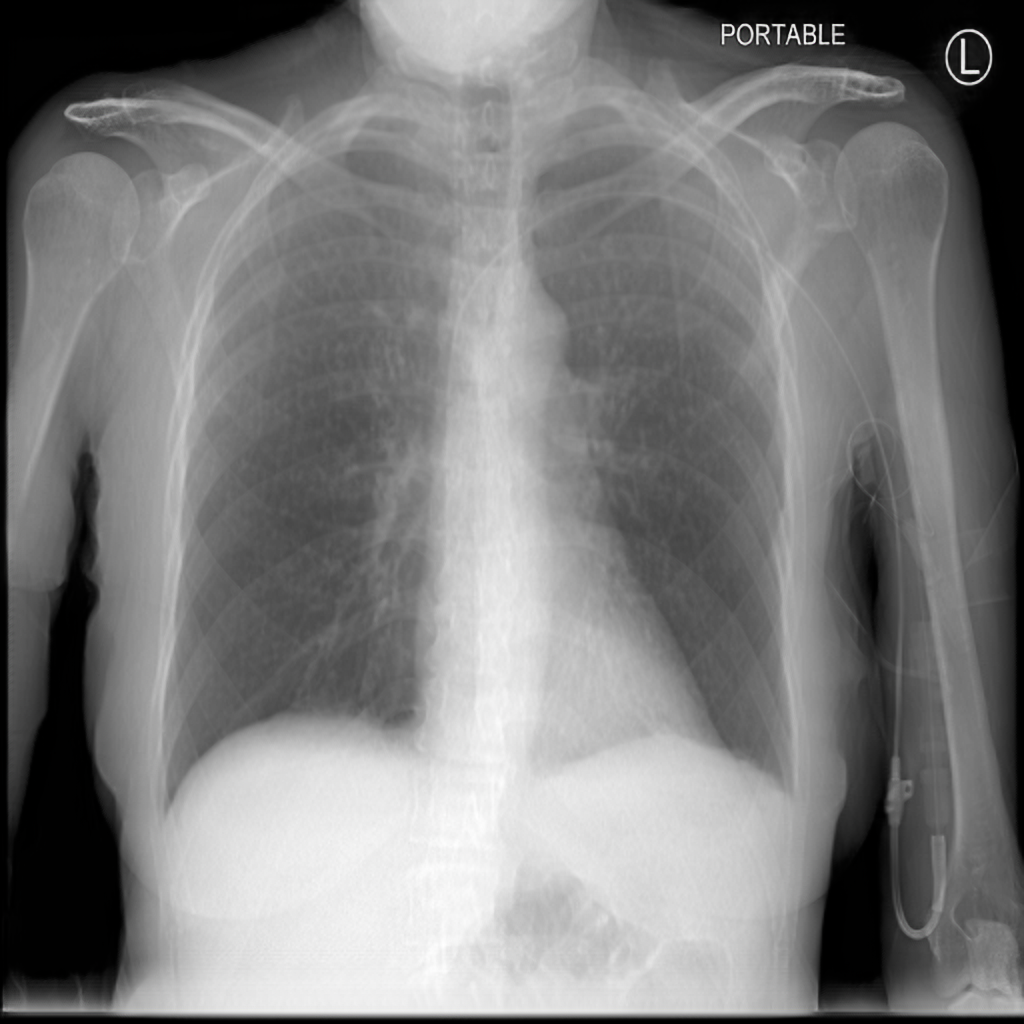}} &
{\includegraphics[width=0.2\textwidth]{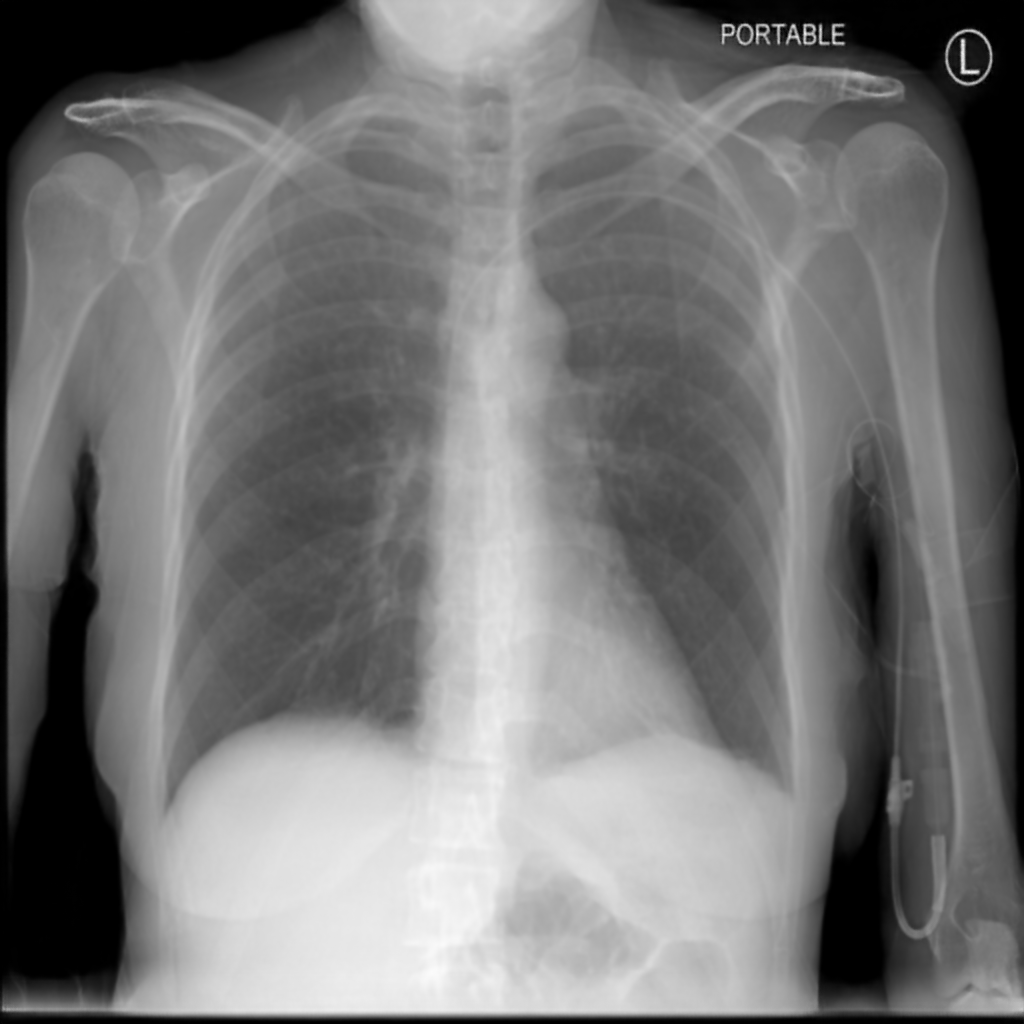}} \\
% Row 3
{\includegraphics[width=0.2\textwidth]{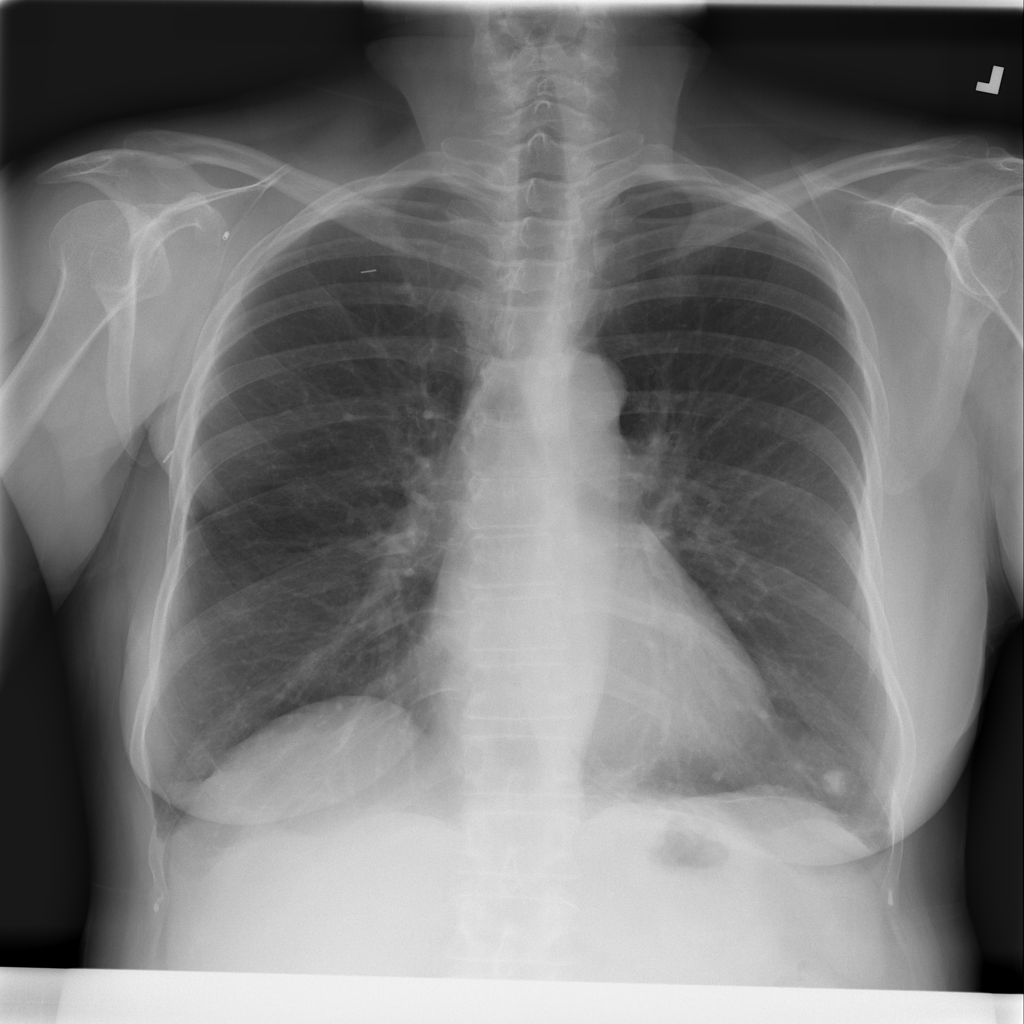}}&
{\includegraphics[width=0.2\textwidth]{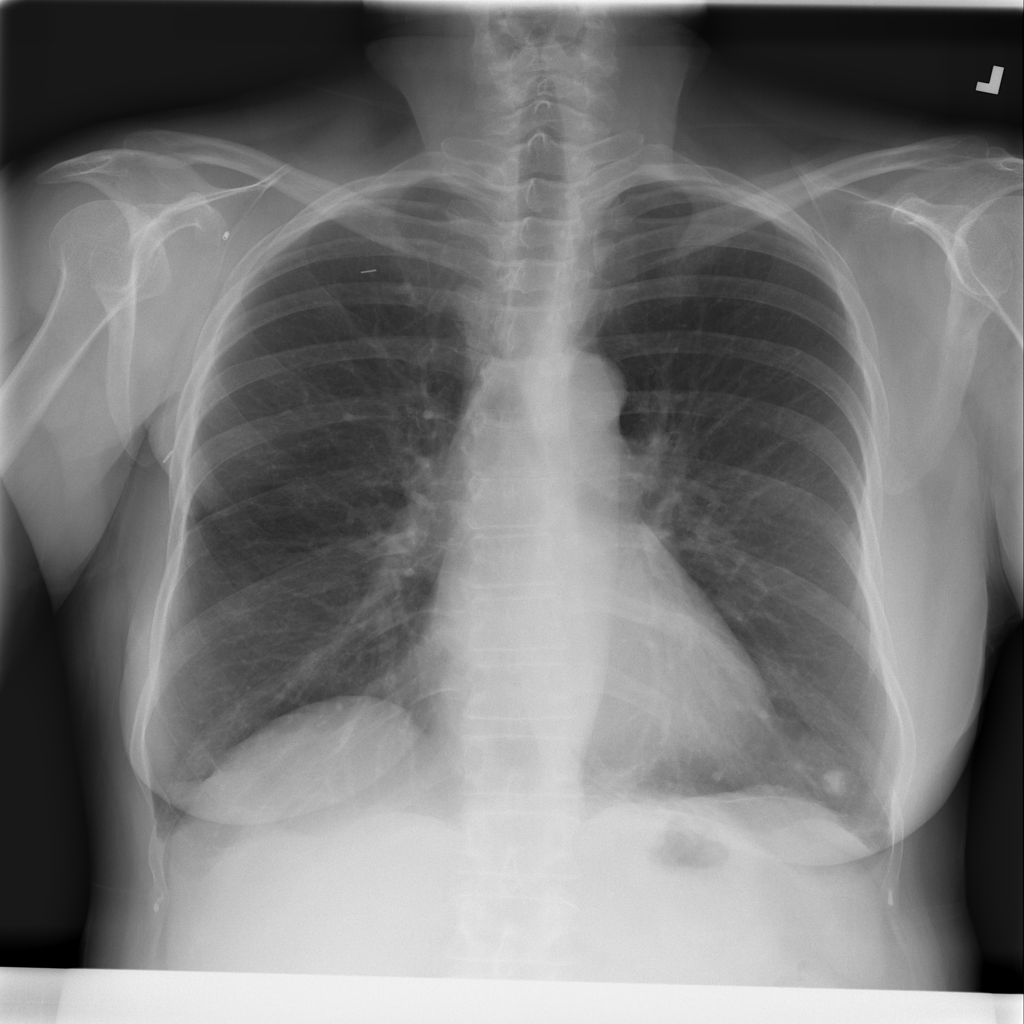}}&
{\includegraphics[width=0.2\textwidth]{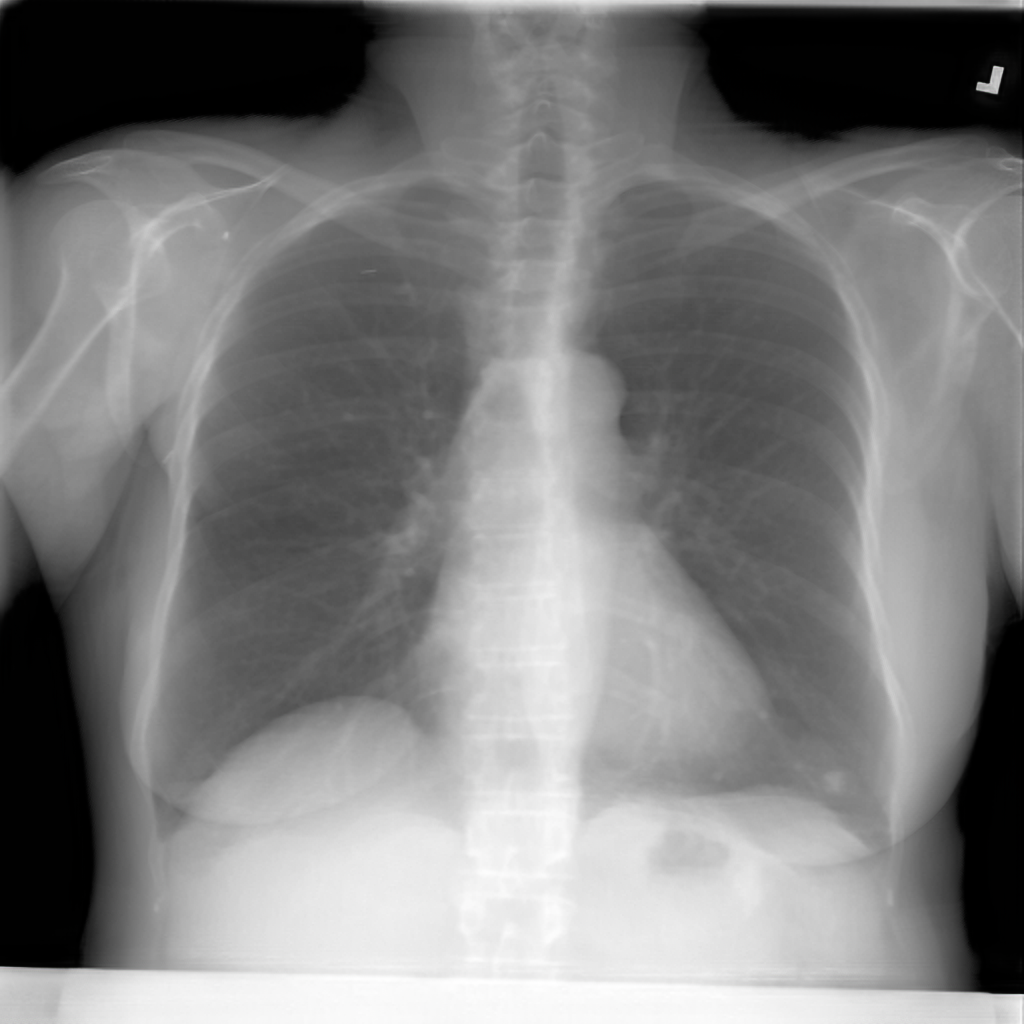}} &
{\includegraphics[width=0.2\textwidth]{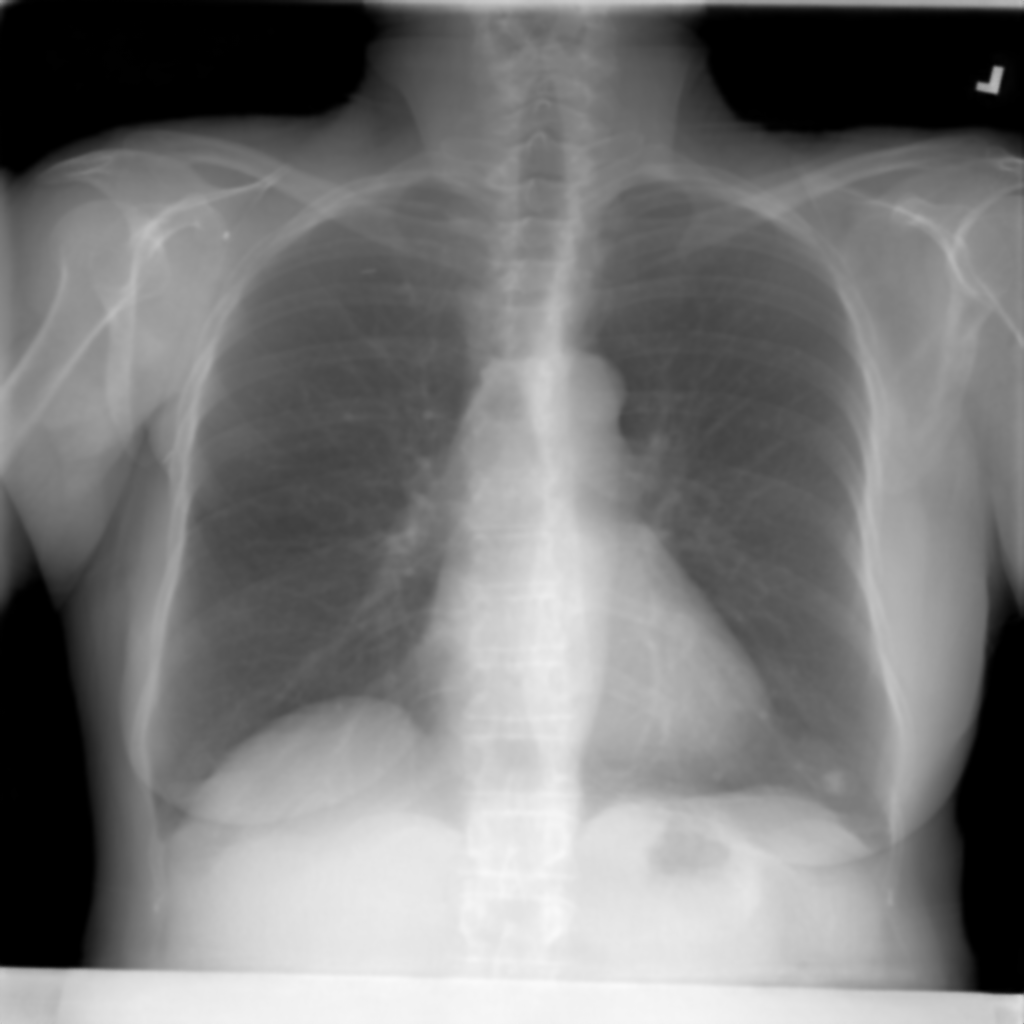}} \\
\end{tabular}
\end{center}
\vspace*{-0.55cm}
\caption{ \small Qualitative comparison of low-quality CXR synthesis from high-quality CXRs using bicubic interpolation, MUNIT-LQ, and DDPM-LQ. \vspace*{-.4cm}
}

\label{fig:lq_generation_comparison}
\end{figure*}

\section{Conclusion}
% In this work, We present a novel image enhancement pipeline - DiffusionXRay, designed to enhance DRRs produced from CT Scans specifically for cases with subtle features prone to distortion or loss of contrast, demonstrated on chest X-ray data. We achieve the ability of image enhancement by first forming realistic training pairs which was identified as the bottleneck in training of enhancement model via MUNIT-LQ and DDPM-LQ models.
% Qualitative and Quantitative evaluations showcase improvements and validate it to be a good alternative for data augmentation but a major setback being computationally expensive. Otherwise, Future work could address various conditioning mechanisms, particularly focusing on selective feature preservation while optimizing computational efficiency.

We introduce DiffusionXRay, an image enhancement pipeline for digitally reconstructed radiographs (DRRs) from CT scans, addressing the challenge of subtle feature distortion and contrast loss. By constructing realistic training pairs using MUNIT-LQ and DDPM-LQ, we overcome a key bottleneck in enhancement model training.

Qualitative and quantitative evaluations confirm significant improvements, validating DiffusionXRay as a viable data augmentation tool. However, its high computational cost remains a limitation, which future work could mitigate by improving efficiency and exploring adaptive conditioning for selective feature preservation.
Beyond DRRs, DiffusionXRay can be adapted to other tasks such as portable X-rays enhancement. Future research could refine the approach to enhance X-rays across varying acquisition conditions while maintaining diagnostic integrity.

\noindent\textbf{Disclosure of Interests:} The authors have no competing interests to declare that are relevant to the content of this paper.

\bibliographystyle{splncs04}
\bibliography{refs}

\end{document}